# A New Reliable & Parsimonious Learning Strategy Comprising Two Layers of Gaussian Processes, to Address Inhomogeneous Empirical Correlation Structures

**Gargi Roy**            Gargi.Roy@brunel.ac.uk
**Dalia Chakrabarty**            Dalia.Chakrabarty@brunel.ac.uk
Department of Mathematics, Brunel University London
Middlesex, Uxbridge UB8 3PH, United Kingdom

## Abstract

We present a new strategy for learning the functional relation between a pair of variables, while addressing inhomogeneities in the correlation structure of the available data, by modelling the sought function as a sample function of a non-stationary Gaussian Process (GP), that nests within itself multiple other GPs, each of which we prove can be stationary, thereby establishing sufficiency of two GP layers. In fact, a non-stationary kernel is envisaged, with each hyperparameter set as dependent on the sample function drawn from the outer non-stationary GP, such that a new sample function is drawn at every pair of input values at which the kernel is computed. However, such a model cannot be implemented, and we substitute this by recalling that the average effect of drawing different sample functions from a given GP is equivalent to that of drawing a sample function from each of a set of GPs that are rendered different, as updated during the equilibrium stage of the undertaken inference (via MCMC). The kernel is fully non-parametric, and it suffices to learn one hyperparameter per layer of GP, for each dimension of the input variable. We illustrate this new learning strategy on a real dataset.

## 1 Introduction

Learning the generic non-linear relationship between two variables, stands challenged in the face of small-to-medium-sized data that is marked by heterogeneities in its correlation structure. Eschewing model-driven supervised learning, a fully non-parametric learning strategy is seeded in the treatment of the sought function as a random variable, the probability distribution of which is then given by an appropriately chosen – generally non-stationary – stochastic process. The desired fast and (preferably) closed-form predictions are potentially promised under such a treatment. The endeavour then reduces to the learning of the correlation function of the underlying stochastic process, such that (s.t.) it replicates (the heterogeneous) correlation structure of the data. It is then also desired that such correlation-transference be computationally easy, achieved under a well-defined algorithm, and involve the learning of only a few parameters (Noack et al., 2023). In this paper we present a learning strategy that is compatible with such demands and desirables, with the generative process chosen as a Gaussian Process (GP) (Rasmussen and Williams (2006); Schulz et al. (2018)), owing to high generalisability across dimensions; ease of computation; and importantly, since a GP minimally constraints sample functions drawn from it.

Much work has been undertaken in the past on usage of non-stationary kernels, (Sampson and Guttorp, 1992; Barry et al., 1996; Higdon, 1998; Schmidt and O'Hagan, 2003; Bornn et al., 2012). The need for nonstationarity in the basal stochastic process (aka a GP), given inhomogeneities in the correlation structures of typical real-world datasets, has led to work on non-stationary kernels.

In this context, Higdon et al. (1999) developed a non-stationary kernel, based on the convolution of the kernels centred at the input locations. Using this work as foundation, Paciorek and Schervish (2003) have proposed a non-stationary correlation function that can be built from a stationary correlation (Paciorek (2003)). Adams and Stegle (2008) introduced the input-dependent variability of the amplitude of the covariance kernel, using the product of two stationary Gaussian processes. Ba and Joseph (2012) achieve non-stationary behaviour by summing two stationary GPs and Davis et al. (2021) extend this model to a Bayesian one.

Heinonen et al. (2016) modelled the length scale; signal variance; and noise variance, with latent GPs within a two-layered model, to achieve nonstationarity, where the latent GPs are ascribed squared exponential covariance kernels. Remes et al. (2017) proposed a non-stationary spectral kernel that again invokes two layers of GPs in which each hyperparameter of the kernel is modelled with a latent GP. In fact, they have used the Gibbs non-stationary kernel (Gibbs, 1998) where the length scale is modelled as an input-dependent function. A two-layered learning structure is also used by (Paciorek and Schervish, 2003).

A common problem with such two-layered strategies is that the GPs in the inner layer are all considered stationary, but no justification for this assumption is advanced. Now we do not generally possess any information in any new application, on values of the hyperparameters of the kernels that parametrise the covariance function of the GPs in the inner layer. Thus, suggestions (as those made in the literature) that these hyperparameters "can be set based on prior knowledge or using grid-search over suitable values", do not solve the problem given that a protocol for such a suggested search is not available to assess computational complexity of the learning. More to the point, there typically exists no prior information on kernel hyperparameters - such as the length scale that one needs to move by along a direction in input space, for inter-output correlation to fade by a given factor. We may possess information on the prior probability distribution of the output at a given input, but that does not inform on the mapping from the space of a kernel hyperparameter to the space of the output.

In the non-stationary spectral mixture kernel reported by Remes et al. (2017) $Q$ number of mixture components are used. The number of hyperparameters to be learnt depends on $Q$ and the number of input dimensions. Paciorek and Schervish (2003) construct a non-stationary covariance function, the input-dependence of which is manifest in a pair of (generally) matrix-valued parameters defined at the pair of inputs, at which the kernel is being computed – to parametrise the correlation between the output variables that are realised at these inputs. The evolution of these kernel matrices determines how frequently the covariance structure changes in the input space. For a $d$-dimensional input, the number of hyperparameters that need to be learnt is of the order of $\mathcal{O}(d^2)$. The complexity of the parameterisation technique discourages this method for $d > 5$. Thus, the existing two-layered strategies appear to challenge our overarching need to learn few parameters. One is also left wondering why in these models, further layers are not added to model the hyperparameters of the GPs in the inner layer. Thus, correctness of the model, as well as parsimony of the strategy need to be understood.

In this paper we present a new non-parametric kernel based on two layers of GPs, with a non-stationary GP in the outer layer that has $d$ number of stationary GPs nested within it – where such stationarity is proved – with $d$ being the number of hyperparameters of the covariance kernel used to parametrise the covariance function of the outer GP. The sufficiency of the two layers is also proven. The kernel is shown to be equivalently non-stationary, and is at the same time, maximally-parsimonious, s.t. the learning of a scalar-valued univariate function requires the learning of one hyperparameter for each layer of the GP. We demonstrate the method on a real-world training set that demands a highly non-linear temporal variation of the output variable. We make comparisons against predictions made by different versions of the method, as well as to results on the same data obtained using multiple non-stationary and stationary kernels that exist in the literature. We show that predictive power of our non-parametric kernel is higher than that of such extant non-stationary kernels. We also present comparisons against Deep Neural Networks (DNNs) to demonstrate superiority of our predictions, and well-definedness of our model.

The paper is organised as follows. Section 2 introduces our fully non-parametric model with the relevant preliminaries. Thereafter, Section 3 discusses the construction of the non-stationary equivalent of this non-parametric model. Aspects of our non-parametric and non-stationary models are blended to create a third model still – we refer to this as the "Blended model" (in Section 4). Then Section 5 presents empirical illustrations of the presented models, along with the comparisons against results of extant models, using a real world dataset. Finally we conclude this paper in Section 6.

## 2 Model

We learn the relation between an input variable $X \in \mathcal{X} \in \mathbb{R}$ and the output variable $Y \in \mathcal{Y} \in \mathbb{R}$, where this relation is depicted as: $Y = f(X)$. Indeed, in this paper, we focus on the learning of a univariate, scalar-valued function $f(\cdot)$. We will be extending this methodology to learn a high-dimensional function in a future contribution. We learn $f: \mathcal{X} \longrightarrow \mathcal{Y}$, using the available training dataset $\mathbf{D}_{train}$ that consists of the set of $M$ pairs of input-output values: $\mathbf{D}_{train} := \{(x_i, y_i)\}_{i=1}^{M}$.

The sought function $f(\cdot)$ is an unknown, and as with any unknown in the Bayesian paradigm, $f(\cdot)$ is a random variable in our model. We will then ascribe a probability distribution to this function-valued random variable; such a distribution is then given by a stochastic process. In other words, we treat $f(\cdot)$ as a sample function of an adequately-chosen stochastic process. We choose this process to be a Gaussian Process (GP), given that a GP imposes minimal constraints on the sought function; is easily generalised across dimensions; and is easy to compute. We refer to this GP as $\mathcal{GP}_{outer}$.

**Definition 1** $f(\cdot) \sim \mathcal{GP}_{outer}(\mu(\cdot), cov(\cdot, \cdot))$, where $\mu(\cdot)$ and $cov(\cdot, \cdot)$ are the mean and covariance functions of $\mathcal{GP}_{outer}$. Then joint probability of a finite number of realisations of $f(\cdot)$ - such as $M$ realisations of $f(\cdot)$ at the $M$ design points $x_1, \cdots, x_M$ - is

$$[f(x_1), f(x_2), \ldots, f(x_M)] = MN(\boldsymbol{\mu}, \boldsymbol{\Sigma}),$$

where $MN(\boldsymbol{\mu}, \boldsymbol{\Sigma})$ is the Multivariate Normal density with an $M$-dimensional mean vector $\boldsymbol{\mu}$ and an $M \times M$-dimensional covariance matrix $\boldsymbol{\Sigma}^{(M \times M)} = [cov(Y_i, Y_j)]$, where $Y_i$ is the random variable $f(x_i)$ realised at $X = x_i$, $\forall i = 1, \ldots, M$.

**Remark 2** We standardise data on the output using sample mean $\bar{y} = \sum_{i=1}^{M} y_i/M$ and sample variance $\sum_{i=1}^{M}(y_i - \bar{y})^2/(M-1)$, and undertake the learning of $f(\cdot)$ with the standardised data. Predictions made with the thus learnt $f(\cdot)$ are unstandardised using the same value of sample mean and of the sample variance. This implies $\boldsymbol{\mu} = \underline{0}$ and $\boldsymbol{\Sigma}$ is a correlation matrix that we denote as: $\boldsymbol{\Sigma} = [corr(Y_i, Y_j)]$.

**Definition 3** We undertake kernel-parametrisation of the covariance function of $\mathcal{GP}_{outer}$, s.t. correlation matrix $\boldsymbol{\Sigma}$ is then equivalently $[corr(Y_i, Y_j)] = [K(x_i, x_j; \theta_{ij}^{(1)}, \theta_{ij}^{(2)}, \ldots, \theta_{ij}^{(H)})]$. Here the kernel $K(\cdot, \cdot)$ is a declining function of a "difference" measure between the inputs $x_i$ an $x_j$; the kernel computed at these inputs bears $H$ hyperparameters $\theta_{ij}^{(1)}, \ldots, \theta_{ij}^{(H)}$, s.t. $\boldsymbol{\theta}_{ij} = (\theta_{ij}^{(1)}, \ldots, \theta_{ij}^{(H)})^T$. Here $\theta_{ij}^{(k)} \in \mathbb{R}$ $\forall k = 1, \ldots, H$, and $i, j = 1, \ldots, M$.

In this definition of $\boldsymbol{\Sigma}$, we set the kernel hyperparameters to depend on the inputs to the kernel, in anticipation of location-dependence (within input space $\mathcal{X}$) of the correlation between pairs of output variables that are realised at a given design input pair. In other words, we invoke a non-stationary GP to model $f(\cdot)$, given training set $\mathbf{D}_{train} = \{(x_i, y_i)\}_{i=1}^{M}$, where output variable $Y_i$ is realised as $y_i$ at the $i$-th design input, i.e. at $X = x_i$.

**Remark 4** $\theta_{ij}^{(k)} = \alpha_k(x_i, x_j)$ - for a random function $\alpha_k : \mathcal{X} \times \mathcal{X} \longrightarrow \mathbb{R}$ - implies $\mathcal{GP}_{outer}$ is non-stationary, $\forall k = 1, \ldots, H$. However, this way of modelling nonstationarity does not lead to a computationally feasible learning exercise, since in this naïve parametric non-stationary model, we would need to learn $\sim M(M-1)/2$ number of kernel hyperparameters at each of the distinct $M(M-1)/2$ input pairs that mark the training data $\mathbf{D}_{train}$ at hand. Thus, such an undertaking involves the undesirable overhead of learning a huge number of hyperparameters. Hence it is infeasible to use a model in which kernel hyperparameters are made to vary from one input pair to another.

Following on, we take a different approach to include nonstationarity in the model for the kernel hyperparameters. Firstly, we view there to be $H$ kernel hyperparameters: $\theta_1, \ldots, \theta_H$. We model the $k$-th kernel hyperparameter $\theta_k$ to be an unknown function $\phi_k(\cdot, \cdot)$ of the relevant inputs, s.t. at $X = x$ and $X = x^/$, $\theta_k = \phi_k(x, x^/)$.

In our Bayesian approach, $\phi_k(\cdot)$ is a random function-valued variable, modelled as a sample function of a GP called $\mathcal{GP}_k$, s.t. at input $X = x$,

$$\theta_k = \phi_k(x) \text{ where } \phi_k \sim \mathcal{GP}_k(\mu_k(\cdot), \kappa_k(\cdot, \cdot)),$$

where $\mu_k(\cdot)$ and $\kappa_k(\cdot, \cdot)$ are mean and covariance functions of $\mathcal{GP}_k$. This holds for $k = 1, \ldots, H$.

## 2.1 Kernel hyperparameters vary with sample function of $\mathcal{GP}_{outer}$

**Remark 5** *A change to the shape of the sample function - that is drawn from $\mathcal{GP}_{outer}$ - implies changes to those parameters that affect inter-output correlation, (such as the length scale over which we need to move in input space for such correlation to fade by an identified value). In other words, when the shape of the sample function $\tilde{f}(x)$ (from $\mathcal{GP}_{outer}$) changes, the kernel hyperparameters $\theta_1, \ldots, \theta_H$ change. Then it follows that we model $\theta_k(x_i, x_j)$ as a function $\phi_k(\cdot)$ of the (shape of, and therefore of the sample function) $\tilde{f}(\cdot)$ and $x_i, x_j$. Indeed, as the shape of the sample function $\tilde{f}$ changes, any kernel hyperparameter computed at the input pair $x_i, x_j$ is then affected by the change in $\tilde{f}(x)$ - which we then consider modelling as $\theta_k(x_i, x_j) = \phi_k(\tilde{f}(x), x_i, x_j)$. However, in our model we consider the dependence on the input $X$ via the sample function $\tilde{f}(\cdot)$ s.t. we model $\theta_k(x_i, x_j) = \phi_k(\tilde{f}^{(i,j)}(x))$, where a new realisation $\tilde{f}^{(i,j)}(\cdot)$ of the sample function $\tilde{f}(\cdot)$ is drawn from $\mathcal{GP}_{outer}$ at each $x_i, x_j$ pair. This proposed model for $\theta_k$ holds for $k = 1, \ldots, H$. This is therefore a non-stationary model.*

We will undertake a fully non-parametric learning of the relation between input $X$ and output $Y$, s.t. the $X - Y$ relationship $f(\cdot)$ is non-parametric, with the underlying correlation structure – parametrised by kernel hyperparameters $\theta_1, \ldots, \theta_H$ - is also learnt non-parametrically, as distinguished from situations in which the kernel is specified with an *ad hoc* parametric form. Then implementation of the above proposed model calls for $\phi_k(\cdot)$ to be learnt using another GP – denoted $\mathcal{GP}_k$ - $\forall k = 1, \ldots, H$.

However, it is not feasible to implement such a model. It may be perceived that this is because the learning uses a function-valued input, though the problem is deeper. Indeed one approach to ease the difficulty with a function-valued input is to discretise the domain of the sample function into $N_{bins}$ number of partitions, and hold $\tilde{f}(\cdot)$ a constant over each such partition, s.t. $\tilde{f}(\cdot)$ is replaced by an $N_{bins}$-dimensional vector. Then the training data for the learning of $\phi_k(\cdot)$ under this scheme will demand the pair: this vectorised-version of $\tilde{f}(x)$ where $\tilde{f}(x)$ is drawn from $\mathcal{GP}_{outer}$ that is specified by a chosen $\theta_k$ - and this chosen $\theta_k$. Then the bigger problem alluded to above is that we would need to draw a different realisation of the sample function at each input pair, and therefore, the training set will be distinct at each input pair, and will also be rendered different every time the random function-valued input changes – such as at distinct iterations of the inferential strategy that will need to be undertaken. Therefore, we need to use the proposed model - that will capture nonstationarity of $\mathcal{GP}_{outer}$ very well – to produce a different model that is implementable (and efficient).

## 2.2 Best model for kernel hyperparameters

**Remark 6** *The best model for $\theta_k$ is advanced under the suggestion that the system - that generates values of the output variable at designed values of the input variable $X$ - is ergodic. Then the average output at $X = x$, given a set $\{\tilde{f}^{(\cdot,\cdot)_j}(\cdot)\}_{j=1}^{N}$ of realisations of the sample function $\tilde{f}(\cdot)$ is the same, irrespective of whether $N$ distinct samples are randomly drawn from a given non-stationary GP, or whether a sample is drawn from each of $N$ distinct non-stationary GPs - each of which is specified by a distinct kernel-parametrised covariance function. The kernel hyperparameters of any of these $N$ GPs are updated during one of $N$ iterations within the iterative learning strategy, (or more precisely, the iterative inference) that is undertaken to learn unknowns. Here $N$ is a large number.*

Let the iteration index variable be $T \in \mathcal{T} \subseteq \mathbb{Z}$.

**Example 1** *If we undertake Bayesian inference using Markov Chain Monte Carlo methods (MCMC) (Robert and Casella (2005); Andrieu et al. (2003)) then in each of $N$ iterations in the equilibrated part of an MCMC chain, the kernel hyperparameters are updated, and with the updated kernel, the covariance function of $\mathcal{GP}_{outer}$ is updated. From this updated process, a sample function is drawn. Thus, once such an MCMC chain - that facilitates such an iterative learning of $f(\cdot)$ - is allowed to run for long s.t. it has equilibrated and is ergodic, samples drawn at varying iterations, result in the same average output as the average obtained using multiple random samples drawn in any iteration from a given non-stationary GP.*

**Lemma 7** *Modelling $\theta_k = \phi_k(\tilde{f}^{(\cdot,\cdot)}(x))$ for sample function $\tilde{f}(\cdot)$ of $\mathcal{GP}_{outer}$, where random function $\phi_k : \mathcal{Y} \longrightarrow \mathbb{R}$, is equivalent to modelling $\theta_k = q_k(t)$ at time-step (or iteration) index $T = t$, for random function $q_k : \mathcal{T} \subset \mathbb{Z} \longrightarrow \mathbb{R}$, $\forall t > n_0$, where by the $n_0$-th time step, the system that outputs $Y = y_i$ at design point $X = x_i$ is ergodic. Thus, $n_0$ is large. Here $k = 1, \ldots, H$.*

Thus, the impractical but ideal model for $\theta_k$ that we discussed in the previous subsection, is equivalent to the comparatively easier-to-implement model:

$$\theta_k = \phi_k(\tilde{f}^{(\cdot,\cdot)}(x)) \iff \theta_k = q_k(t),$$

$t = 0, 1, 2, \ldots, N_{iter}$; $k = 1, \ldots, H$, after running the iterative (inference) for the large number $(n_0)$ of iterations.

**Definition 8** *The unknown function $q_k(\cdot)$ is a function-valued random variable that we model as a sample function of a GP that we denote $\mathcal{GP}_k$:*

$$q_k(\cdot) \sim \mathcal{GP}_k(\mu_k(\cdot), \kappa_k(\cdot, \cdot)).$$

Here $\mu_k(\cdot)$ and $\kappa_k(\cdot, \cdot)$ are the mean and covariance functions of $\mathcal{GP}_k$ which is the $k$-th GP in the inner layer, for $k = 1, \ldots, H$.

**Remark 9** *$\mathcal{GP}_{outer}$ comprises the outer layer of our learning strategy, s.t. the sought function $f(\cdot)$ is modelled with a sample function of $\mathcal{GP}_{outer}$. The $k$-th hyperparameter of the covariance function of $\mathcal{GP}_{outer}$ is modelled as a function $q_k()$ of the time-step variable $T$, with $q_k(\cdot)$ modelled with a sample function of $\mathcal{GP}_k$, $\forall k = 1, \ldots, H$.*

## 2.3 Inner layer can be built with stationary GPs

**Theorem 10** *Let a hyperparameter of the kernel that parametrises the covariance function $Cov(\cdot, \cdot)$ of $\mathcal{GP}_{outer}$, be $\theta_k \in \mathbb{R}$, $\forall k = 1, \ldots, H$. We model hyperparameter $\theta_k$ as $\theta_k = q_k(t)$, where $t$ is a realisation of the time-step parameter $T \in \mathcal{T} \subset \mathbb{N}$, s.t. $q_k : \mathcal{T} \subset \mathbb{N} \longrightarrow \mathbb{R}$ is Lipschitz continuous.*

This is a standard result that states continuity of any mapping from the space of natural numbers to $\mathbb{R}$.

**Proof** A distance $d_{i,j}$ between the time step inputs $T = t_i$ and $T = t_j$ to function $q_k(\cdot)$ is $|t_i-t_j|$. Here $t_i, t_j \in \mathcal{T} \subset \mathbb{N}$.

A distance $d(q_k(t_i), q_k(t_j))$ between values of the output $\theta_k \in \mathbb{R}$ computed at inputs $t_i$ and $t_j$ is $|q_k(t_i) - q_k(t_j)|$.

Assume that for a finite $\alpha > 0$,

$$\frac{d(q_k(t_i), q_k(t_j))}{\alpha} > d_{i,j}, \quad \forall t_1, t_2 \in \mathcal{T}.$$

Then it follows from this assumption that for $\dfrac{d(q_k(t_i), q_k(t_j))}{\alpha} = 1/2$, $|t_i - t_j| < 1/2$. However, $|t_i - t_j| < 1/2$ implies $t_i = t_j \ \forall i, j$. This is false. Therefore our assumption is wrong, i.e.

$$\frac{d(q_k(t_i), q_k(t_j))}{\alpha} \leq d_{i,j}, \quad \forall t_1, t_2 \in \mathcal{T}.$$

Hence, $q_k : \mathcal{T} \subset \mathbb{N} \longrightarrow \mathbb{R}$ is Lipschitz continuous. ∎

**Theorem 11** *Correlation function of a (weakly) stationary process $\{X_t\}_{t \in \mathcal{T}}$ is continuous.*

**Proof** We consider $\{X_t\}_{t \in \mathcal{T}}$ to be a zero-mean process. Let $\delta \in \mathbb{R}$. Then recalling that for this weakly stationary process $\mathbb{E}(X_t) = \mathbb{E}(X_{t+\delta})$, and that each expectation is 0 given that the process is a zero-mean process, $|Corr(X_{t+\delta}, X_0) - Corr(X_t, X_0)|^2 = |\mathbb{E}(X_{t+\delta} X_0) - \mathbb{E}(X_t X_0)|^2 = |\mathbb{E}((X_{t+\delta} - X_t)X_0)|^2 \leq \mathbb{E}(X_{t+\delta} - X_t)^2 \mathbb{E}(X_0)^2$, where the last inequality stems from Cauchy Schwartz.

But, $\mathbb{E}(X_{t+\delta} - X_t)^2 = \mathbb{E}(X_{t+\delta})^2 + \mathbb{E}(X_t)^2 - 2\mathbb{E}(X_{t+\delta} X_t)$, where $\mathbb{E}(X_t)^2 = Var(X_t) = \mathbb{E}(X_0)^2$ $\forall t$ in this stationary process.

Using all this, $|Corr(X_{t+\delta}, X_0) - Corr(X_t, X_0)|^2 \leq \mathbb{E}(X_\delta)^2 + \mathbb{E}(X_0)^2 - 2\mathbb{E}(X_0 X_\delta)$, i.e.

$$|Corr(X_{t+\delta}, X_0) - Corr(X_t, X_0)|^2 \leq (\mathbb{E}(X_\delta) - \mathbb{E}(X_0))^2.$$

Then for this process, as $\delta \longrightarrow 0$, $|Corr(X_{t+\delta}, X_0) - Corr(X_t, X_0)|^2 \longrightarrow 0$, i.e. the correlation function of this weakly stationary process, is continuous $\forall t \in \mathcal{T}$. ∎

**Theorem 12** *The GP $\mathcal{GP}_k$ - sample functions of which models the function $q_k(\cdot)$ - can be a stationary process with a continuous correlation function, $\forall k = 1, \ldots, H$.*

**Proof** By Theorem 10, $q_k(\cdot)$ is a Lipschitz continuous function, $\forall k = 1, \ldots, H$ at all inputs $t = N_{burnin}, N_{burnin}+1, \ldots, N_{burnin}+N_{iter}$. Here $T$ is the time-step of an iterative inferential exercise, run for $N_{iter}$ iterations, where the iterative inference attains equilibrium at the $N_{burnin}$-th iteration.

By Theorem 1 of Cambanis (1973), in a stationary GP that has a continuous correlation function, either all sample functions are continuous, or all sample functions are unbounded in every finite interval.

Here $q_k(\cdot)$ is modelled as a sample function drawn from $\mathcal{GP}_k$, and $q_k : \mathcal{T} \longrightarrow \mathbb{R}$ is a Lipschitz continuous map, i.e. no sample functions of $\mathcal{GP}_k$ is unbounded in any finite interval.

$\implies \mathcal{GP}_k$ if modelled as a stationary process, will generate $q_k(\cdot)$ correctly, as long as we can show that $\mathcal{GP}_k$ has a continuous correlation function.

But by Theorem 11, correlation function of a stationary $\mathcal{GP}_k$ is continuous.

Thus, modelling $\mathcal{GP}_k$ as a stationary process - which has a continuous correlation function - will correctly generate sample functions that are not unbounded in any finite interval, as is true of

$q_k : \mathcal{T} \longrightarrow \mathbb{R}$.
Hence it suffices to model $\mathcal{GP}_k$ as stationary. This holds $\forall k = 1, \ldots, H$. ∎

**Remark 13** *Indeed, $\mathcal{GP}_k$ is not proven to necessarily be stationary, but it suffices for it to be stationary to underlie the observation that all sample functions drawn from $\mathcal{GP}_k$ are continuous.*

Thus, we will hold $\mathcal{GP}_k$ to be stationary $\forall k = 1, \ldots, H$. Therefore all hyperparameters of $\mathcal{GP}_k$ will be treated as constants, that we will learn given the training data relevant to the learning of $q_k : \mathcal{T} \subset \mathbb{N} \longrightarrow \mathbb{R}$.

### 2.4 Training data required to learn $q_k(\cdot)$

Given the model $\theta_k = q_k(t)$, learning $q_k(\cdot)$ cannot be undertaken with data on $X$ and $Y$. The training data relevant for this learning will need to comprise pairs of values of: iteration index $T$, and the updated value of $\theta_k$ in the $t$-th iteration. To the best of our knowledge, the need for such data as distinct from the empirical data – where such data is relevant at the inner layer, for the learning of the location-dependent kernel hyperparameters – has not been reported. This is typically because the GP ($\mathcal{GP}_k$) that $q_k(\cdot)$ is generated from, is pre-assigned a chosen covariance function, (Paciorek and Schervish, 2003; Heinonen et al., 2016; Remes et al., 2017), unlike our method in which we learn the covariance of the GPs in the inner layer, ($\forall k = 1, \ldots, H$). Even under the assumption that the GPs in the inner layer are stationary – where justification for this very assumption is missing in the literature – the (constant) kernel hyperparameters of the covariance structures of such GPs need to be learnt.

Using the empirical data $\mathbf{D}_{train}$ alone, implies that learning of the sought $f(\cdot)$ is undertaken via percolation of information in $\mathbf{D}_{train}$ into this inner layer via the kernel hyperparameters that are invoked within the likelihood, where such kernel hyperparameters are set as $X$-dependent, with this $X$-dependence modelled as a sample function of a GP. This approach indeed does not offer any scope for scanning across the covariance structure of any of the $H$ GPs that exist within the inner layer. Thus, lack of adequate training data at the inner layer compels kernel hyperparameters of $\mathcal{GP}_1, \ldots, \mathcal{GP}_H$ to be held constants, and not learnt in the data. Such is however not satisfactory since the aim of advancing a general methodology that caters to pan-disciplinary applications, stands challenged when parameters of all GPs in the inner layer are fixed by hand.

**Definition 14** *To learn $q_k(\cdot)$, we formulate training dataset at the current, i.e. the t-th iteration. This data is $\mathbf{D}_k^{(t)} := \{(i, \theta_k^{(i)})\}_{i=t-1}^{t-N_{LB}}$, where $\theta_k^{(i)}$ is the current value of $\theta_k$ in the i-th iteration of the MCMC chain run to learn the sought function $f(\cdot)$, as well as $q_1(\cdot) \ldots, q_{H(\cdot)}$. Here $k = 1, \ldots, H$. Then $i \leq t$, where $t = N_{LB} + 1, \ldots, N_{iter}$ for a chosen width $N_{LB}$ of time interval in the immediate past, within which we look back, to collate known values of $\theta_k$. Here $N_{iter}$ the total number of iterations of this MCMC chain.*

**Remark 15** *Thus, $\mathbf{D}_k^{(t)}$ is the dynamically-varying look-back data that bears inputs from the past $N_{LB}$ learnt values of $\ell$.*

### 2.5 "SQE-looking" and "Matérn-looking" kernels used

For standardised data on the outputs $Y_1, \ldots, Y_M$ - where $Y_i$ is realised as $y_i$, at the $i$-th design point $x_i$ in $\mathbf{D}_{train}$ - the correlation matrix $\boldsymbol{\Sigma}$ of the Multivariate Normal likelihood induced by $\mathcal{GP}_{outer}$, is $\boldsymbol{\Sigma} = [corr(Y_i, Y_j) = [K(x_i, x_j)]; \theta_1, \ldots \theta_H)]$, with basal shape of kernel $K(x, x^{/})$ chosen as "SQE"-looking or "Matérn"-looking in our work.

**Definition 16**

$$SQE\text{-looking kernel}: \quad K(x, x'; \ell(\cdot)) = \exp\left(-\frac{(x-x')^2}{2\ell(x)\ell(x')}\right);$$

$$\textit{Matérn-looking kernel}:$$

$$K(x, x'; \ell(\cdot), \nu(\cdot)) = \frac{2^{1-\nu(\cdot)}}{\Gamma(\nu(\cdot))}\left(\frac{\sqrt{2\nu(\cdot)}}{\ell(\cdot)}d(x, x')\right)^{\nu(\cdot)} K_\nu\left(\frac{\sqrt{2\nu(\cdot)}}{\ell(\cdot)}d(x, x')\right),$$

where $\theta_k = q_k(t)$ and $q_k(\cdot) \sim \mathcal{GP}_k(\mu_k(\cdot), \kappa_k(\cdot, \cdot))$, $\forall k = 1, \ldots, H$.

When we invoke an SQE-looking kernel, $H = 1$, i.e. we learn a single (length scale) hyperparameter $\ell \equiv \theta_1$ in the first block of any iteration given the (static) empirically-observed data on $X, Y$ pairs. In the second block we learn the hyperparameter $\delta$ of the single $\mathcal{GP}$ in the inner layer, as relevant to this simple learning application. Then our method entails the learning of only 2 parameters in the case that the sought function $f(\cdot)$ is univariate and scalar-valued.

When we invoke the Matérn-looking kernel, $H = 2$ since we learn both the length scale hyperparameter $\ell \equiv \theta_1$ and the roughness hyperparameter $\nu \equiv \theta_2$. Then each hyperparameter is treated as a distinct random function ($q_1(\cdot)$ and $q_2(\cdot)$ respectively) of the time-step or iteration index $T$, with $q_k(\cdot) \sim \mathcal{GP}_k(\mu_k(\cdot), \kappa_k(\cdot, \cdot))$, for $k = 1, 2$. Since $\kappa_k(\cdot, \cdot)$ is parametrised using a stationary SQE kernel, we learn a single hyperparameter ($\delta_k$) of each GP in the inner layer. Thus, we then learn 4 hyperparameters in total.

## 2.6 Inference

We have used MCMC-based inference in our work.

- In the first block of the $t$-th iteration,

  1. likelihood of $\theta_1, \ldots, \ell_H$ in the data $\mathbf{D}_{train}$ is multivariate Normal:

  $$\mathcal{L}(\theta_1, \ldots, \theta_H | \mathbf{D}) = \frac{1}{\sqrt{(2\pi)^M |\mathbf{\Sigma}|}} \exp\left(-\frac{1}{2}(\boldsymbol{Y} - \boldsymbol{\mu})^T \mathbf{\Sigma}^{-1}(\boldsymbol{Y} - \boldsymbol{\mu})\right), \quad (1)$$

  where the standardised values of outputs realised at the design inputs are collated into the vector $\boldsymbol{Y} = (y_1, \ldots, y_M)^T$; $\boldsymbol{\mu} = \mathbf{0}$; and $\mathbf{\Sigma} = [corr(Y, Y')] = [K(x, x', \theta_1, \ldots, \theta_H)]$, with $K(\cdot, \cdot; \theta_1, \ldots, \theta_H)$ either SQE-looking, or Matérn-looking in our work, (see Definition 16).

  2. In the $t$-th iteration, we propose value $\theta_k^{(t)}$ of $\theta_k$ as: $\theta_k^{(\star,t)} \sim \mathcal{N}(\theta_k^{(t-1)}, s_k^2)$, where a Normal proposal density is used for $\theta_k \in \mathbb{R}$. On the other hand, if $\theta_k$ is non-negative, it is proposed from a truncated Normal density. $\theta_k^{(t-1)}$ is the value of $\theta_k$ current at the end of the $t-1$-th iteration; proposal variance $s_k^2$ is experimentally chosen.

  3. Likelihood of hyperparameters $\theta_1, \ldots, \theta_H$ is computed at these proposed values, as well as their respective current values - in the data $\mathbf{D}_{train}$. This leads to the computation of the joint posterior $\pi(\theta_1, \ldots, \theta_H | \mathbf{D}_{train})$, at their current and proposed values, after invoking relevant priors on $\theta_1, \ldots, \theta_H$.

  4. Acceptance ratio $\alpha_t$ is computed at $t$-th iteration:

  $$\alpha_t = \frac{\pi(\theta_1^{(\star,t)}, \ldots, \theta_H^{(\star,t)} | \mathbf{D}_{train}) \prod_{k=1}^{H} p_k(\theta_k^{(t-1)}, s_k^2)}{\pi(\theta_1^{(t-1)}, \ldots, \theta_H^{(t-1)} | \mathbf{D}_{train}) \prod_{k=1}^{H} p_k(\theta_k^{(\star,t)}, s_k^2)},$$

  where $p_k(\cdot, s_k^2)$ is the proposal density for $\theta_k$, (computed at proposed and current value of $\theta_k$).

5. If $\alpha_t \geq u$, where $U \sim Uniform[0,1]$ it implies $\theta_k^{(t)} = \theta_k^{(\star,t)}$; else $\theta_k^{(t)} = \theta_k^{(t-1)}$, $\forall k = 1, \ldots, H$.

6. Here, the updating of $\theta_k$ is done as relevant to the end of the first block of the $t$-th iteration - to be revised at the end of the second block of the $t$-th iteration, if $t > n_0$, where $n_0 > N_{LB}$. If $0 < t \leq n_0$, there is only one block in the $t$-th iteration, and the $\theta_k$ just updated in the $t$-th iteration serves as the current value of $\theta_k$ in this iteration.

- 1. In the second block of the $t$-th iteration - for $t > N_{LB}$ - we start by recalling that $\theta_k = q_k(\cdot)$, with $q_k(\cdot) \sim \mathcal{GP}_k(\mu_k, \kappa_k(\cdot, \cdot; \delta_k))$ for $k = 1, \ldots, H$.

2. We standardise the lookback data: $\mathbf{D}_k^{(t)} = \{(i, \theta_k^{(i)})\}_{i=t-N_{LB}}^{t-1}$, s.t. likelihood of $\delta_k$ in data $\mathbf{D}_k^{(t)}$ is $MN(\boldsymbol{\mu}_k^{(t)}, \boldsymbol{\Delta}_k^{(t)})$, $\boldsymbol{\mu}_k^{(t)} = \mathbf{0}$ and correlation matrix $\boldsymbol{\Delta}_k^{(t)} = [\exp(-(t_i - t_j)^2/2\delta_k^2)]$. This holds for $k = 1, \ldots, H$.

3. In the second block of the $t$-th iteration, we propose $\delta_k^{(\star,t)} \sim \mathcal{N}(\delta_k^{(t-1)}, \psi_k^2)$, $\forall k = 1, \ldots, H$. We compute the likelihood of $\theta_1, \ldots, \theta_H$ in the data $\mathbf{D}_1^{(t)}, \ldots, \mathbf{D}_H^{(t)}$ respectively, at the proposed and current values of these hyperparameter variables, leading to posterior $\pi(\delta_k^{(\star,t)}|\mathbf{D}_k^{(t)})$ of $\delta_k$ given the relevant lookback data.

4. Thereby we compute the acceptance ratio $\alpha_t^{(2nd)} := \dfrac{\prod\limits_{k=1}^{H} \pi(\delta_k^{(\star,t)}|\mathbf{D}_k^{(t)})}{\prod\limits_{k=1}^{H} \pi(\delta_k^{(t-1)}|\mathbf{D}_k^{(t)})}$, s.t. if $\alpha_t^{(2nd)} \geq u'$, for $U = u'$, $\delta_k^{(t)} = \delta_k^{(\star,t)}$; else $\delta_k^{(t)} = \delta_k^{(\star,t-1)}$, $\forall k = 1, \ldots, H$. The updated value of $\theta_k$ at the end of the second block of the $t$-th iteration defines the updated $\boldsymbol{\Delta}_k^{(t)}$ matrix.

5. Ultimately, we employ this updated $\boldsymbol{\Delta}_k^{(t)}$ matrix to predict the expectation of $\theta_k$ - which henceforth will be treated as the current value $\theta_k^{(t)}$ of $\theta_k$ in the $t$-the iteration. This predicted mean value is:

$$\theta_k^{(t)} = (e^{-((t-1)-t)^2/2(\delta_k^{(t)})^2} \ldots e^{-(t-N_{LB}-t)^2/2(\delta_k^{(t)})^2})\boldsymbol{\Delta}_k^{(t)}(\theta_k^{(t-1)}, \ldots, \theta_k^{(t-N_{LB})})^T,$$

where $\boldsymbol{\Delta}_k^{(t)} = [\exp(-(i-j)/2(\delta_k^{(t)})^2]$, for $i, j = t - N_{LB}, \ldots, t-1$.

For our SQE-looking implementation, $H = 1$ and $\theta_1 \equiv \ell$ and there is a single hyperparameter $\delta$ (of the SQE kernel that is invoked to parametrise the covariance function of the stationary GP in the inner layer). We propose $\ell$ in the first block of the $t$-th iteration from $\mathcal{N}(\ell^{(t-1)}, s_\ell^2)$, $\forall t = 1, \ldots, N_{iter}$ and update $\ell$ using the accept-reject criterion of the Random Walk Metropolis Hastings that is undertaken within the first block of our Metropolis-within-Gibbs inferential strategy. In the second block of the same iteration we perform the following: we first build the lookback data $\mathbf{D}^{(t)}$ for $t > N_{LB}$. Then we propose $\delta$ from $\mathcal{N}(\delta^{(t-1)}, s_\delta^2)$. By inserting the posterior of $\delta$ given $\mathbf{D}^{(t)}$ - computed at the proposed and current values of $\delta$ - into the acceptance criterion of the undertaken Random Walk Metropolis Hastings algorithm, we update $\delta$. Using the updated $\delta$ and $\mathbf{D}^{(t)} = \{(i, \ell^{(i)})\}_{i=t-N_{LB}}^{t-1}$, we predict the value of $\ell$ at the end of the second block of the $t$-th iteration. This predicted $\ell$ is now the current value of $\ell$, as of the end of the $t$-th iteration.

For the Matérn-looking implementation, $H = 2$ and $\theta_1 \equiv \ell$; $\theta_2 \equiv \nu$. Again, there are two hyperparameters - $\delta_1, \delta_2$ - of the SQE kernels used to specify the stationary GPs in the inner layer. In the second block of the $t$-th iteration – for $t > N_{LB}$ - $\delta_1$ is proposed from $\mathcal{N}(\delta_1^{(t-1)}, s_{\delta_1}^2)$; $\delta_2$ is proposed from $\mathcal{N}(\delta_2^{(t-1)}, s_{\delta_2}^2)$.

All through, we use Gaussian priors on $\ell$ - and $\nu$, when relevant. Then the unscaled posterior of the hyperparameter(s) is given by the product of this likelihood and prior. The priors on the hyperparameters are experimentally chosen.

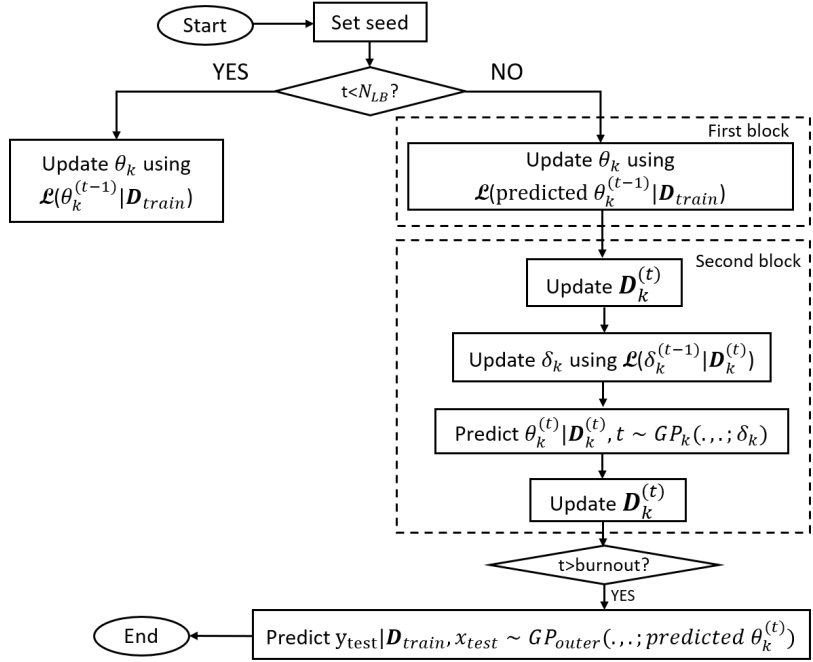

Figure 1: Figure depicting flow of the dual-layered GP-based learning strategy that uses a non-parametric kernel.

This learning scheme is depicted with a flow diagram in Figure 1 and as an algorithm in Algorithm 1.

## 2.7 Single GP layer based models

While we have motivated a fully non-parametric model above, it is also possible to include a single stationary GP to generate sample functions that will be used to model the sought function $f(\cdot)$. Such a stationary kernel could be an SQE or a Matérn - the one or two (respectively) constant hyperparameters of which we will learn, given the training dataset. For a literature on kernels see Genton (2001), on automatically designing new stationary kernels composed from base kernels see Duvenaud et al. (2013), on stationary spectral mixture kernel see Wilson and Adams (2013) - though the computational burden of automatically designing new stationary kernels limits the complexity of the selected kernels (Tobar et al. (2015)) and in spite of the expressivity of the spectral kernels, Simpson et al. (2021) suggests that the spectral kernels get limited by its optimisation procedure because the form of such kernel includes multiple modes in the marginal likelihood which poses special challenges to the optimisation. We will display results from SQE and Matérn stationary single-layer models to showcase their shortcomings when modelling a real-world functional relationship between $X$ and $Y$.

In this model the sought function $f(\cdot)$ is modelled as the random realisation from a Gaussian process whose covariance structure has been modelled using a parametric, stationary kernel. Therefore, $f \sim \mathcal{GP}(\mu(x), cov(x, x'))$, where $\mu(\cdot)$ and $cov(\cdot, \cdot)$ are mean and covariance functions of this GP. Then by definition, the joint probability of a finite number of realisations of $f(\cdot)$ - such as $M$ realisations of $f(\cdot)$ at $M$ design points $x_1, \cdots, x_M$ in the training data $\mathbf{D}_{train}$ - is Multivariate Normal, with the $M$-dimensional mean vector $\boldsymbol{\mu}$ and covariance matrix $\boldsymbol{\Sigma}^{(M \times M)}$.

We parametrise matrix $\boldsymbol{\Sigma}$ by saying that the $ij$-th element of this matrix that represents the covariance between the variable $Y_i$ - that represents the output at $X = x_i$ - and $Y_j$, is modelled as

**Algorithm 1:** Implementation of the non-parametric kernel.

```
1  ∀k = 1, ..., H
2  At t=0, set seed for θ_k
3  for t ← 1 to N_iter increment by 1 do
4      if t < N_LB then
5          Update θ_k using L(θ_1, ..., θ_H | D_train)
6      else
7          if t = N_LB then
               /* Block-1                                                      */
8              Update θ_k using L(θ_1, ..., θ_H | D_train)
               /* Block-2                                                      */
9              Construct D_k^(t) = {{i}_{i=t-N_LB}^{t-1}, {θ_k^(t-N_LB), ..., θ_k^(t-1)}}
10             Update δ_k using L(δ_k | D_k^(t))
11             Predict θ_k^(t) | D_k^(t), t ~ GP_k(μ_k(·), κ_k(.,.; δ_k))
12             Modify D_k^(t) as insert θ_k^(t) at end, remove first item, s.t. D_k^(t) =
                   {{i}_{i=t-N_LB}^{t-1}, {θ_k^(t-N_LB+1), ..., θ_k^(t-1), θ_k^(t)}}
13         else
               /* Block-1                                                      */
14             Update θ_k as θ_k^(t) using predicted θ_k^(t-1) as current value, L(θ_1, ..., θ_H | D_train)
               /* Block-2                                                      */
15             Modify D_k^(t) as insert θ_k^(t) at end, remove first data point
16             Repeat line 10, 11 and 15 in order
17     Post burnin, predict at input test points using predicted θ_k^(t) as:
           f(x*) | D_train, x* ~ GP_outer(μ(·), cov(·, ·; (θ_1^(t), ..., θ_H^(t))))
```

a covariance function $K_S(\cdot, \cdot)$ of the distance between the inputs $x_i$ and $x_j$ at which these output variables are realised. But the joint of these $M$ output variables, is the probability of the data on $Y$. In fact, it is a conditional probability - conditional on the parameters of the mean and the covariance matrix. Hence this multivariate Normal becomes the likelihood of the inferential scheme. We will present results of this Single GP layer based parametric model where stationary SQE and Matérn kernel is used as $K_S(\cdot, \cdot)$ respectively.

## 3 Non-Stationary extension of the non-parametric kernel

In Section 2.1 we clarified that the sample function dependent model of a kernel hyperparameter is infeasible, though it is inherently non-stationary and includes the required information to correctly formulate $\theta_k$. Then we invoked equivalence of sampling across time-step and sampling across input pairs, to suggest the *non-parametric model*, in which $\theta_k = q_k(t)$ with $q_k(\cdot) \sim \mathcal{GP}_k(\mu_k(\cdot), \kappa_k(\cdot, \cdot))$ where $\mathcal{GP}_k$ proven to be stationary, $\forall k = 1, \ldots, H$.

However, we are also interested in learning the hyperparameters - such as the length scales of the correlation structure - at each input pair. To this effect, in this section we establish the non-stationary equivalent of the non-parametric kernel. We construct this non-stationary kernel as an SQE-looking kernel with length scale hyperparameter $\ell_{i,j}$ that is relevant to input pair $X = x_i$ and $X = x_j$, where $x_i$ and $x_j$ are the $i$-th and $j$-th design points in training set $\mathbf{D}_{train}$. We implement this non-stationary kernel to parametrise $corr(Y_i, Y_j), \forall i, j = 1, \ldots, M$.

**Remark 17** *To construct the non-stationary equivalent of the non-parametric kernel advanced above, we first construct a sample of realisations of the random variable $Y_i$ that is output at $X = x_i$, and another sample of realisations of $Y_j$ output at $X = x_j$. The two samples can then be used to estimate $corr(Y_i, Y_j)$ - as $\hat{corr}(Y_i, Y_j)$ - which we will then equate to the SQE-looking non-stationary kernel: $K(x_i, x_j; \ell_{i,j})$, (to compute $\ell_{i,j}$, $\forall ij = 1, \ldots, M$).*

Thus, we are not attempting the learning of the (generally very large number, i.e. the) $M(M-1)/2$ number of such input-pair dependent hyperparameters – a task that we have ruled as infeasible early in Section 2 - but we are simply computing $\ell_{i,j}$ to forward a non-stationary equivalent of the non-parametric kernel that is already learnt. We will then make predictions at a set of test inputs using the non-stationary and non-parametric kernel models, using real-world data bearing an inhomogeneous correlation, to demonstrate said equivalence. However, from an implementation point of view, we will employ the non-parametric kernel model, which once established as equivalent to a non-stationary model, is understood as capable of real-world applications, and is easier to implement.

We start constructing the non-stationary kernel, after we have finished running the MCMC chain to make inference on parameters of $\mathcal{GP}_{outer}$ and $\mathcal{GP}_1, \ldots, \mathcal{GP}_H$, using the SQE-looking non-parametric kernel. The generation of the sample $S_i$ of random realisations of $Y_i$ is first undertaken.

**Remark 18** *Since $x_i$ is a design point, all predictions of $Y_i$ will fundamentally be $y_i$, where we recall the training set to be $\mathbf{D} = \{(x_i, y_i)\}_{i=1}^{M}$. Then instead of sampling from the posterior predictive of $Y_i$ multiple times, we sample at multiple values of $X$ that live within an $\epsilon$-neighbourhood of $X = x_i$, in the input space $\mathcal{X}$. Here $\epsilon > 0$ and any "neighbour" of $x_i$ is at distance $\leq \epsilon$ from $x_i$ in $\mathcal{X}$.*

**Definition 19** *$N_s$ neighbours of $x_i$ in $\mathcal{X}$ are: $x_i^{(1)}, x_i^{(2)}, \ldots, x_i^{(N_s)}$, where $|x_i^{(u)} - x_i| \leq \epsilon$, $\forall u = 1, 2, \ldots, N_s$.*

**Theorem 20** *Correlation matrix $\mathbf{\Sigma}$ of the multivariate Normal likelihood induced by $\mathcal{GP}_{outer}$ is parametrised as: $\mathbf{\Sigma} = [corr(Y_i, Y_j)] = [\exp(-(x_i - x_j)^2 / 2\ell_{i,j}^2)]$, with*

$$\ell_{i,j} = \sqrt{\frac{-(x_i - x_j)^2}{\ln(\hat{corr}(Y_i, Y_j))}}.$$

*Here we employ an SQE-looking kernel for $\mathcal{GP}_{outer}$ and $Y_i$ is the output realised at $X = x_i$, $i = 1, \ldots, M$.*

**Proof** Using the non-parametric model in which only one hyperparameter $\theta$ - modelled as $\theta = q(t)$ - is relevant to the (SQE-looking) kernel employed to parametrise the covariance function of $\mathcal{GP}_{outer}$, we predict the expected output value $y_i^{(p)}$ of the output at $X = x_i^{(p)}$, at iteration index $T = t$ as:

$$y_i^{(p)} | \mathbf{D}, x_i^{(p)} = (e^{-((x_i^{(p)} - x_1)^2 / 2q(t)^2)} \ldots e^{-(x_i^{(p)} - x_M)^2 / 2q(t)^2})(\mathbf{\Sigma})^{-1}(y_1, \ldots, y_M)^T,$$

where the correlation matrix $\mathbf{\Sigma}_i^{(p)} = [\exp(-(x_i - x_j)^2 / 2(q(t))^2)]$, in the non-parametric kernel model; $i, j = 1, \ldots, M$. Then we define sample

$$S_i = \{y_i^{(1)}, y_i^{(2)}, \ldots, y_i^{(N_s)}\}, \forall i = 1, \ldots, M.$$

Then the unbiased estimated correlation $\hat{corr}(Y_i, Y_j)$ using samples $S_i$ and $S_j$ is:

$$\hat{corr}(Y_i, Y_j) = \frac{\sum_{p=1}^{N_s}((y_i^{(p)} - \bar{y}_i)(y_j^{(p)} - \bar{y}_j))}{(N_s - 1)},$$

where, $\bar{y}_i$ and $\bar{y}_j$ are sample mean of $S_i$ and $S_j$ respectively.

Then we equate $\hat{corr}(Y_i, Y_j)$ to the (SQE-looking) non-stationary kernel-parametrised correlation:

$$\hat{corr}(Y_i, Y_j) = \exp\left(-\frac{(x_i - x_j)^2}{(\ell_{i,j})^2}\right),$$

s.t.

$$\ell_{i,j} = \sqrt{\frac{-(x_i - x_j)^2}{\ln(\hat{corr}(Y_i, Y_j))}}.$$

This holds $\forall i \neq j$, $i, j = 1, \ldots, M$. For $i = j$, the length scale of the correlation function is unity. ∎

In fact, we will also compute such length scales at pairs of test inputs – by generating neighbours within a distance $\epsilon$ from a given test input and predicting output values at such neighbours. Additionally, proceeding along same lines, we also compute length scale hyperparameters at a pair of inputs, one of which is a test point, while the other is a design input. Such length scales are invoked in prediction of the expectation and variance of the output $Y_{test}$ realised at a test input at $X = x_{test}$:

$$\mathbb{E}(Y_{test}|\mathbf{D}, x_{test}) = (e^{-((x_{test}-x_1)^2/(\ell_{test,1})^2} \ldots e^{-(x_{test}-x_M)^2/(\ell_{test,M})^2})\mathbf{\Sigma}^{-1}(y_1, \ldots, y_M)^T, \quad (2)$$

$$Var(Y_{test})|\mathbf{D}, x_{test} = 1 - (e^{-((x_{test}-x_1)^2/(\ell_{test,1})^2} \ldots e^{-(x_{test}-x_M)^2/(\ell_{test,M})^2})\mathbf{\Sigma}^{-1}$$
$$(e^{-((x_1-x_{test})^2/(\ell_{test,1})^2} \ldots e^{-(x_M-x_{test})^2/(\ell_{test,M})^2}). \quad (3)$$

(We recall that $\mathbf{\Sigma}$ is rendered the correlation matrix upon standardisation of output values).

If the underlying non-parametric kernel is not SQE-looking, but is marked by hyperparameters $\theta_1, \ldots, \theta_H$, we can construct the SQE-looking non-stationary equivalent kernel using the same protocol as discussed above. Then elements of the sample $S_i$ will be outputs predicted at chosen $\epsilon$-neighbours of $x_i$ using this non-SQE-looking non-parametric kernel that now hosts multiple hyperparameters – this is the only difference to the above discussion. However, the non-stationary kernel construction begins after all hyperparameters of the non-parametric kernel are learnt. So we will be able to predict such outputs $\forall i = 1, \ldots, M$.

For a vector-valued input, we will undertake this construction by generating neighbours of a given point in $\mathcal{X}$ within a distance $\epsilon$, along each direction in $\mathcal{X}$, and predictions of outputs made at each neighbour along the $k$-th such direction will contribute towards learning the length scales pertaining to the $k$-th direction, for all directions relevant for the $d$-dimensional input variable. We undertake such an application in a future contribution.

### 3.1 Shortcomings of the non-stationary kernel model

Estimation of $\{\ell_{i,j}\}_{i<j; i,j=1}^M$ can suffer from numerical instability. Since we are using the SQE-looking basal kernel – of which $\ell$ is the length scale hyperparameter (that takes a distinct value at a given input pair) – a small fractional error in computation of $\hat{corr}(Y_i, Y_j)$ can imply a large fractional error in $\ell_{i,j}$.

**Remark 21** *For the SQE-looking kernel, fractional error in $\ell_{ij}$ is $-1/\ln((\hat{corr}(Y_i, Y_j))^2)$ times the fractional error in the estimated correlation between $Y_i$ an $Y_j$.*

In particular, the construction of the non-stationary kernel implies that $corr(Y_i, Y_j)$ is approximated by the empirical estimate of correlation computed using elements in samples $S_i$ and $S_j$ that comprise output values at $\epsilon$-neighbours of $x_i$ and $x_j$. However, if sample functions of $\mathcal{GP}_{outer}$ are highly rough within an $\epsilon$-neighbourhood of $x_i$, the empirical correlation estimate may not approximate $corr(Y_i, Y_j)$. The computed $\ell_{i,j}$ can then be rendered incorrect, especially given the sensitive dependence of $\ell_{i,j}$ on $\hat{corr}(Y_i, Y_j)$. All through in this section, $i < j; i, j = 1, 2, \ldots, M$.

We will compare the predictive performance of the non-parametric and non-stationary kernels below. However, given that numerical computation of $\ell_{i,j}$ can be unstable for data manifesting high inhomogeneities in correlation structure, on local scales, we also explore another model that is fed jointly by the non-parametric and the non-stationary kernel models.

## 4 A non-stationary non-parametric kernel

We denote this new kernel model that blends the learnt non-parametric kernel, and its equivalent non-stationary kernel, as the *"blended" kernel model*. We describe it for a scalar-valued input $X$.

Likelihood in this model is defined using declining functions of the absolute difference between the values of the expectation of $Y_i^{(test,A)}$ predicted at test input $x_i^{(test)}$, using the $A$ kernel model, where $A = non-parametric, non-stationary$. In our implementation, we in fact use an SQE-looking non-parametric kernel.

**Definition 22** *In the Blended model, likelihood of $\theta_1, \ldots, \theta_H$ in data $\mathbf{D}_{train}$ is defined as*

$$\mathcal{L}(\theta_1, \ldots, \theta_H; \mathbf{D}_{train}) := \prod_{i=1}^{N_{test}} \exp\left(-\frac{\mathbb{E}(Y_i^{(test,non-parametric)}) - \mathbb{E}(Y_i^{(test,non-stationary)})}{2Var(Y_i^{(test)})}\right),$$

*where $Y_i^{(test,A)}$ is the output variable realised at the i-th test input $x_i^{(test)}$, in the A kernel model, for $A = non-stationary, non-parametric$, $\forall i = 1, \ldots, N_{test}$. Then the mean and variance of $Y_i^{(test,A)}$ are predicted closed-form - see Equation 2 for the expectation and variance predicted at $x_i^{(test)}$.*

We could indeed have chosen any other parametric declining function to define the likelihood, but this does not affect the inferred mean of $\theta_1, \ldots, \theta_H$.

We do not learn the parameters of the inner GPs in this chain; we merely learn $\theta_1, \ldots, \theta_H$.

- We start an MCMC chain in which we learn the non-parametric kernel using the model $\theta_k = q_k(t)$, where $t$ is the value of the iteration index $T$.

- Once the non-parametric kernel is learnt in this chain, each iteration is rendered with a single block. Inside the $t$-th such iteration, we update the hyperparameters $\theta_1, \ldots, \theta_H$ of the non-parametric kernel to $\theta_1^{(t)}, \ldots, \theta_H^{(t)}$. and using this updated non-parametric kernel, we construct an SQE-looking non-stationary kernel with input pair-dependent length scale hyperparameters $\{\ell_{i,j}^{(t)}\}_{i,j=1}^M$ updated a $T = t$.

- We propose values of hyperparameters $\theta_1, \ldots, \theta_H$ of the non-parametric kernel - as $\theta_k^{(\star,t)} \sim p_k(\theta_k^{(t-1)}, s_k^2)$, where $p_k(\cdot, s_k^2)$ is the proposal density that $\theta_k$ is proposed from, for $k = 1, \ldots, H$. This is typically a Normal or truncated Normal density. Here $s_k^2$ is experimentally chosen $\forall k = 1, \ldots, H$.

- In the same iteration, we construct a corresponding non-stationary kernel using this proposed non-parametric kernel. We also compute a non-stationary kernel using the current non-parametric kernel.

- We compute likelihood of this model at the proposed and current values of $\theta_1, \ldots, \theta_H$. Such computation leads to the joint posterior $\pi(\theta_1, \ldots, \theta_H | \mathbf{D}_{train})$ computed at the proposed and current values, after invoking Normal priors of $\theta_1, \ldots, \theta_H$.

- We will accept/reject these proposed values, using an acceptance criterion relevant for the MCMC flavour that we wish to work with.

- At the end of the $t$-th iteration of this chain, we record the predictions on mean of the outputs realised at the test inputs $x_1^{(test)}, x_2^{(test)}, \ldots, x_{N_{test}}^{(test)}$ using the current values $\theta_1^{(t)}, \theta_2^{(t)}, \ldots, \theta_H^{(t)}$, of the hyperparameters of the non-parametric kernel. Thus the predicted mean values of the output at $x_k^{(test)}$ allows for the computation of the relative frequency distribution of this output. We use the histogram constructed for $\theta_k$ - approximating the marginal of $\theta_k$ given the data - to define 95% Highest Probability Density credible region(95% HPDs) learnt on this variable.

## 5 Empirical illustration

We make an empirical illustrations of our developed models - along with comparison of our results with results obtained from other relevant kernels in the literature - using the time series data on Brent crude oil price obtained from https://datasource.kapsarc.org. We choose this dataset as a univariate real world example of a data that exhibits highly non-linear trends in the variation of the output (price) with the input (time). This illustration is on a time series nature of the data, but our model permits diverse applications towards learning of a relationship between a generic output and input pair, to then make fast, uncertainty-included predictions.

### 5.1 Data

The Brent oil prices are publicly available at https://datasource.kapsarc.org for the period: 5th January 2018 to 8th March 2022, though this site does not include data for every day of this interval. From the full data that is available, we choose a time interval - 5th January 2018 to 29th December 2021- to randomly sample 184 data points that constitute the training set $\mathbf{D}_{train}$ that we use in our illustration. We also randomly select 20 time points as test inputs, at each of which we will undertake prediction of oil prices, which we will then compare against the true price. Such test data are precluded from the constructed training set. Thus, the training set $\mathbf{D}_{train}$ in our application comprises 184 training points. We shift the date stamps s.t. the time point of the first data point is denoted the 0-th day; the input to the second time point is then the 1-st day, and so on.

### 5.2 Results of fully non-parametric learning with the non-parametric kernel model

We present results of learning all relevant parameters of the GPs in the outer and inner layers of our fully non-parametric learning. This is done using an SQE-looking and Matérn-looking kernel separately, (as discussed in Section 2.5). Thereafter, we present results of predictions made at the identified test inputs. Inference is performed with Metropolis-within-Gibbs and this MCMC-based inference organically learns the marginal posterior probability density of each learnt parameter, given the training data, resulting the learning of the 95% Highest Probability Density credible regions, (95% HPDs).

**SQE-looking non-parametric kernel:** In our implementation of the SQE-looking non-parametric kernel, from the first block of the Metropolis-within-Gibbs chain, the mean of 95% HPD credible region on the learnt length scale hyperparameter $\ell$ (which is modelled as an unknown function of the iteration index) is learnt to be $\approx 23.94$. The mean $\ell$ value predicted at the end of the second block of post-burnin part the chain is $\approx 23.94$ again. Figure 11 in the Appendix displays the traces of $\ell$; $\delta$; and logarithm of the marginal posterior probability density of each learnt parameter given respective data. Prediction of the output values (oil price values), realised at the test inputs (i.e. test times) is displayed in the left panel of the top row in the Figure 9. We undertook the prediction at the test inputs, at the end of each post-burnin iteration of the conducted Metropolis-within-Gibbs chain; traces of the mean predicted values of the output realisations at four of the test inputs are plotted in Figure 12 of the Appendix.

**Matérn-looking non-parametric kernel:** The chain run with the Matérn-looking non-parametric kernel allowed for the learning of kernel hyperparameters $\ell$ and $\nu$. There are two GPs in the inner layer in this implementation – $\mathcal{GP}_1$ models the relation between iteration index $T$ and $\ell$, while $\mathcal{GP}_2$ models the relation between iteration index and $\nu$. Hyperparameters of the stationary kernel of $\mathcal{GP}_1$ and $\mathcal{GP}_2$ are $\delta_\ell$ and $\delta_\nu$ respectively. The mean of the 95% HPD credible region learnt in the first block of the chain, for the length scale hyperparameter $\ell$ of $\mathcal{GP}_{outer}$, is $\approx 30$. Mean of the $\ell$ value predicted in the second block, is again about 30. Mean learnt value of hyperparameter $\nu$ in the first block of the chain is $\approx 0.97$ and in the second block of the chain it is again 0.97, while mean $\delta_\nu$ is learnt as $\approx 0.3$. Mean of the 95% HPDs learnt on $\delta_\ell$ is $\approx 0.6$. Figure 2 displays the traces for learning of $\ell$ and $\nu$.

Histograms from the traces of output predictions at five test points for fully non-parametric learning with Matérn-looking non-parametric kernel are presented in the Figure 13 of the Appendix. Prediction of oil prices using this implementation of the non-parametric kernel, is depicted in the middle panel of the top row in the Figure 9.

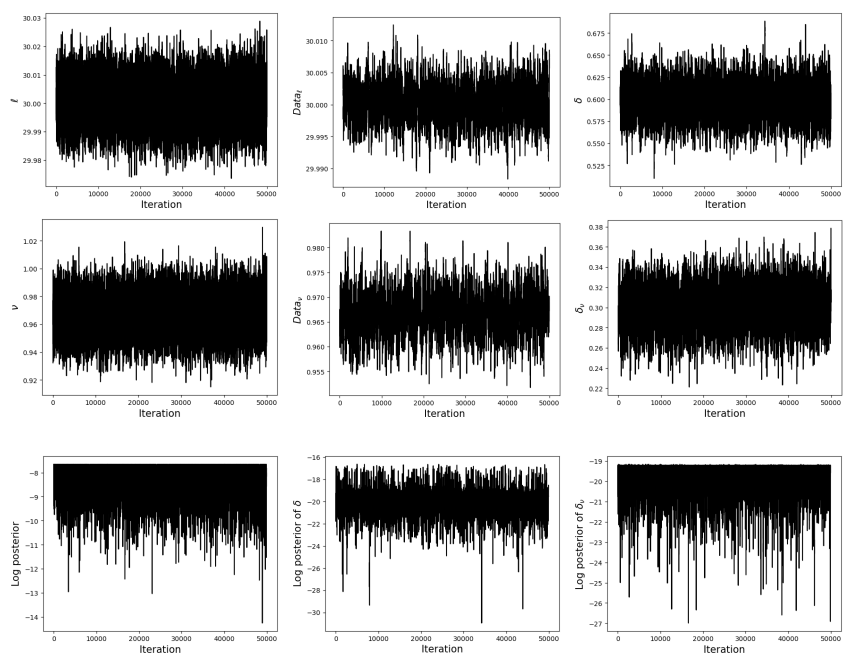

Figure 2: Results of learning $\ell$ and $\nu$ using the non-parametric model with Matérn-looking kernel, given Brent crude oil price data. *Top row, left panel:* trace of $\ell$ updated in the first block of the conducted Metropolis-within-Gibbs chain. *Top centre:* trace of $\ell$ as predicted from the $\mathcal{GP}$ in the inner-layer, at the $\ell$ and $\nu$ that are updated in the first block. *Top right:* trace of $\delta$. *Middle left:* trace of $\nu$ updated in the first block of the chain. *Middle centre:* trace of $\nu$ value predicted using the $\mathcal{GP}$ in the inner-layer. *Middle right:* trace of $\delta_\nu$. *Bottom left:* trace of logarithm of the joint posterior of $\ell$ and $\nu$ given training data. *Bottom middle and right:* traces of logarithm of posterior of $\delta_\ell$ and $\delta_\nu$ respectively, given relevant (dynamically-varying) lookback data.

### 5.3 Results of learning with a non-parametric model, given noisy data

We have undertaken learning with noisy data, where the learning was undertaken with an "SQE looking" non-parametric kernel. Indeed, the output in our work, namely, the Brent crude oil price, is a noise-free variable. So to showcase the capacity in the learning to work with noisy data, we generate said noise. To this effect - assuming Gaussian errors - we superimpose the error variable $\epsilon_e \sim \mathcal{N}(0, \sigma_\epsilon^2)$ on each of the $M$ number of outputs of the training set that is constructed using the Brent crude oil prices, where we choose $\sigma_\epsilon^2 = 0.01$. We refer to the noisy data as $\mathbf{D}^/$. In our learning, we add the matrix $\sigma_\epsilon^2/s^2 \mathbf{I}_M$, to the correlation matrix $\mathbf{\Sigma}$ of the multivariate Normal likelihood depicted in Equation 1. Here, $s$ is the sample standard deviation employed in standardising the outputs of the data; $s \approx 14.57$. Subsequent to the learning with data $\mathbf{D}^/$ and prediction at test inputs, we unstandardise the learnt values of the error variance with $s^2$.

In this learning exercise the error variance $\sigma_\epsilon^2$ is inferred upon in the first block of the Metropolis-within-Gibbs implementation of the MCMC-based inference. From the first block of the Metropolis-within-Gibbs chain run with the constructed noisy data $\mathbf{D}^/$, the mean of 95% HPD credible region on the learnt length scale hyperparameter $\ell$ is learnt to be $\approx 24.02$. The mean of the 95% HPD on predicted $\ell$ value obtained from the second block of the chain is again about 24.01. The mean of the 95% HPD on $\delta$ is learnt as $\approx 0.29$. The learnt mean of the 95% HPD on the error variance $\sigma_\epsilon^2$ is learnt to be about 0.011 (where we recall the true value of this parameter to be 0.01).

Figure 3 displays traces of $\ell$, $\delta$, $\sigma_\epsilon^2$, and of logarithm of the joint posterior of these variables, given the noisy data.

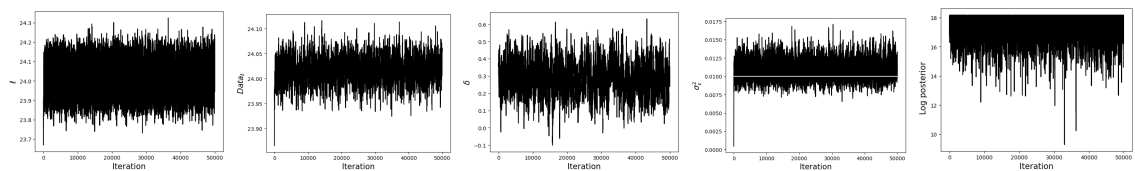

Figure 3: Results of learning with the non-parametric model that uses the SQE-looking kernel, given noisy Brent crude oil price data. The panels from left depict traces of $\ell$; of $\ell$ predicted in second block; of $\delta$; of the error variance $\sigma_\epsilon^2$ (with the true value of $\sigma_\epsilon^2$ depicted in white line); and the logarithm of the joint posterior of $\ell$ and $\sigma_\epsilon^2$, given the data noisy data.

Figure 4 shows histograms of outputs predicted at four test points, where the predictions are undertaken after learning the temporal-dependence of oil price using the SQE-looking non-parametric kernel with noise-free data and with noisy data. The wider of the two histograms (in broken lines), represents the marginal of the predicted output at the chosen test inputs, given the noisy data.

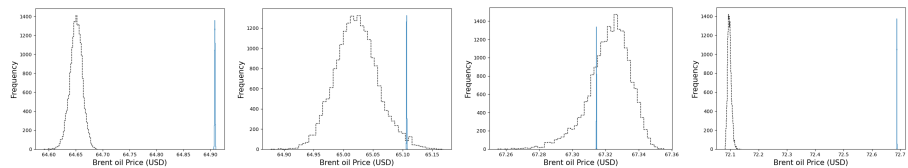

Figure 4: Figure displaying histograms of output values predicted at four test inputs, obtained from the fully non-parametric model - with an SQE-looking kernel that parametrises $\mathcal{GP}_{outer}$ - given noisy data (in black dashed lines), and noise-free data (in blue solid line). The histogram displaying results given the noise-included data, is wider than that depicting prediction results obtained with the noise-free data.

Prediction of the oil prices using this implementation of the non-parametric kernel using noisy oil price, is depicted right panel of the top row in the Figure 9. It is to be noted that the uncertainty in the predictions made given the noisy data (presented in the Figure 4) is in fact about 0.2.

### 5.4 Results of non-stationary model extending from non-parametric model

We have implemented the non-stationary version of the non-parametric model with the SQE-looking kernel as mentioned in Section 5.2. Here, the neighbourhood width $\epsilon$ is set to be 0.092, with $N_s = 100$ neighbouring points constructed within any $\epsilon$-neighbourhood of each input. Prediction of oil prices using this model is depicted in the left panel of the middle row in the Figure 9.

Figure 5 and Figure 6 respectively present heat maps of the correlation matrices and the contour plots of the length scales $\ell_{ij}$ for the learning of the non-stationary equivalent of the non-parametric model that uses an SQE-looking kernel.

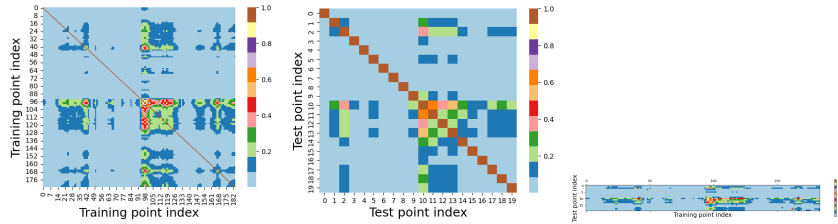

Figure 5: Correlation matrices of the non-stationary equivalent of the non-parametric model that invokes an SQE-looking kernel. The left most plot presents the correlation matrix computed between outputs realised at pairs of design points in the training dataset $\mathbf{D}_{train}$, while the middle plot shows the the correlation matrix computed between outputs realised at input pairs that each comprise test inputs in $D_{test}$. Compared to these panels, the right-most panel depicts the the correlation matrix between an output realised at a design point, and another that is realised at a test input.

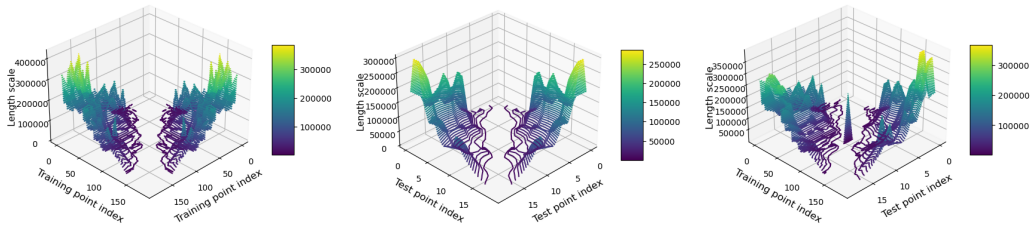

Figure 6: Similar to panels in Figure 5, except here, contour plots of the computed length scales - of the non-stationary equivalent of the non-parametric model - are depicted.

### 5.5 Results of Blended model

In this subsection, we present the results of the Blended model that blends the aspects of the non-parametric learning with SQE-looking kernel and the non-stationary model, hence, $\theta_1 = \ell_B$ is learnt in this Blended model. The mean of the 95% HPD credible region on the learnt length scale hyperparameter, $\ell_B$ in the Blended model is $\approx 30$. Figure 7 presents results of learning in the Blended model. Figure 8 presents traces of few of the input-pair specific length scales $\ell_{ij}$. Traces of the mean output predicted at four test points - which happen to be every fifth test input - are included in Figure 14 of the Appendix. Figure 18 of Appendix 19 present heat maps of the correlation matrices and the contour plots of the length scales $\ell_{ij}$ for this Blended model. Finally, prediction of the oil prices using this model is depicted in the middle panel of the middle row of the Figure 9.

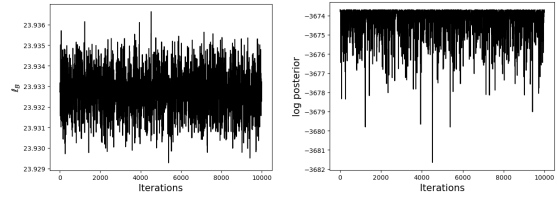

Figure 7: The left panel depicts the trace of the length scale hyperparameter $\ell_B$ of the Blended model. The right panel depicts logarithm of the posterior *pdf* of $\ell_B$ given data $\mathbf{D}_{train}$.

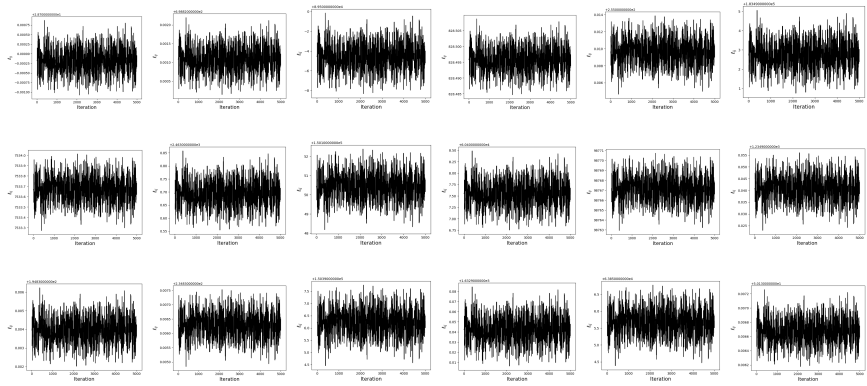

Figure 8: Figure displaying length scale hyperparameters relevant to the implementation of the Blended model. The top row depicts lengths scales computed at input pairs, each of which is a design point. These inputs relevant to the presented plots are (from the left) at the 10-th and 5-th design inputs; the 15-th and the 2nd; the 15-th and the 50-th design points; the 30-th and the 10-th; the 60-th and the 65-th design points; the 180-th and the 10-th. The middle row depicts traces of length scales that are computed at input pairs, each of which is a test input. From the left, panels depict traces of length scale parameters that are relevant to the pair that comprises the 1st and 3rd of the chosen 20 test inputs; the 3rd and 4-th; 2nd and the 14-th test points; the 5-th and the 15-th; the 10-th and the 18-th; and the pair comrising the 19-th and 18-th test points. Bottom row depicts traces of length scales that are computed at input pairs, one of which is a design point, and the other a test input. From the left, the presented traces are pertinent to the pair of the 1st test and the 10-th design inputs; the 1st test and the 25-th design point; the 2nd test and the 100-th design inputs; the 5-th test and the 50-th design; the 9-th test and 25-th design; and the 19-th test and the 180-th design points.

### 5.6 Comparative results from existing stationary and non-stationary kernels

In this subsection we present comparisons of the predictive capacity of multiple existing models, including two well-known stationary kernels; the non-stationary kernel developed by Remes et al. (2017); and that by Paciorek and Schervish (2003).

**Results of single GP layer based model with stationary SQE kernel:** The sole hyperparameter of relevance to this model is the length scale hyperparameter, $\ell$. The mean of its learnt 95% HPD is 4.07. Figure 15 in the Appendix shows the trace of this learnt $\ell$, in addition to the joint log posterior. Predictions of the oil prices (at test times), made using this model are depicted in the right panel of the middle row in the Figure 9.

**Results of single GP layer based model with stationary Matérn kernel:** In this model, the learnt hyperparameters of the kernel are $\ell$ and $\nu$ (depicted in the Equation ). Means of the learnt values of $\ell$ and $\nu$ in the Matérn kernel are 5.29 and 5.39 respectively. Figure 16 in the Appendix displays the learnt traces, in addition to the log posterior. Final predictions using this model are depicted in the left panel of the bottom row in the Figure 9.

| Model | #Hyp. | MSE(mean) | MSE interval | Time |
|---|---|---|---|---|
| Non-Parametric SQE | 2 | 1.2021 | [1.19, 1.21] | 2h:4m:15s |
| Non-Parametric Matérn | 4 | 2.509 | [2.38, 3.19] | 6h:48m:50s |
| Non-Parametric SQE noisy | 3 | 3.7489 | [3.63, 4.03] | 2h:7m:45s |
| Non-Stationary | 2 | 38.6745 | [32.98, 53.02] | 5h:39m:21s |
| Blended | 3 | 38.6731 | [32.97, 53.01] | 13h:31m:8s |
| Stationary SQE | 1 | 52.6213 | [42.62, 67.47] | 1h:48m:2s |
| Stationary Matérn | 2 | 21.8839 | [17.43, 30.03] | 9h:35m:17s |
| Remes et al. (2017) | 3 | 100.694 | [102.79, 224.37] | 4h:1m:46s |
| Paciorek and Schervish (2003) | 1 | 206.5252 | [189.71, 236.48] | 4h:3m:53s |

Table 1: Table displaying comparative predictive performance and computational times relevant to the different models that we have predicted with. The first column depicts the names of the various models. The second column presents the number of hyperparameter(s) learnt using that model, (typically for each dimension of the input vector). The third column depicts the mean squared error (MSE) computed at the predicted means; the fourth column shows the MSE computed using the predicted 95% HPDs; the fifth and right-most column presents the overall time required to compute the predictions. This includes the time to convergence (achieved by the MCMC chain) and predictions made post convergence.

**Results of non-stationary kernel developed by Remes et al. (2017):** One of the extant non-stationary kernels that we learn with, using our training dataset, is a spectral kernel advanced by Remes et al. (2017). Remes et al. (2017) include nonstationarity in their model with a spectral mixture kernel, the number of mixture components in which, can be chosen by hand. However, for each component, three hyperparameters need to be learnt - mixture weight, length scale and frequency, denoted by $w, \ell, \mu$ respectively. In the univariate situation that is relevant to our training set, we have used a single component, to maintain the minimal number of hyperparameters to learn. We have used the likelihood suggested by Remes et al. (2017) and inference with MCMC is undertaken. The mean of the 95% HPD credible region on the learnt hyperparameters, $w$, $\mu$ and $\ell$ in this model are 5.01, 3.8 and 0.29 respectively. Bottom row of the Figure 17 in the Appendix presents the results of learning these hyperparameters. Predictions at test inputs are shown in the middle panel of the bottom row of the Figure 9.

**Results of non-stationary kernel developed by Paciorek and Schervish (2003):** We have also implemented another non-stationary kernel – that developed by Paciorek and Schervish (2003). We have learnt the length scale hyperparameter $\ell$ with a fixed $\nu$ hyperparameter of the Matérn kernel used by Paciorek and Schervish (2003). The value of $\nu$ is fixed at 5.39, as per the parametric model with Matérn kernel that we have employed earlier, in which we have learnt both $\ell, \nu$; (see results in Section 5.6. The mean of the 95% HPD credible region on the learnt $\ell$ is 4.89. Top row of the Figure 17 in the Appendix presents the results of learning with this non-stationary kernel, while predictions from this model are displayed in the right panel of the bottom row in the Figure 9.

### 5.7 Comparative performance for prediction

Table 1 presents the performance of the predictions using all the models that are presented. The performance of the predictions is computed in terms of the mean squared error (MSE) computed for both the predicted mean and the uncertainties i.e. 95% HPD of the predictions for all the models against the true values for the test points. The fully non-parametric model outperforms all other models. Also the non-stationary equivalent of the non-parametric model outperforms non-stationary models from the literature. We also report the time to the computation of predictions, for each model, along with the number of hyperparameters learnt in each model in this univariate empirical illustration undertaken in this paper. All the models are run in a laptop having the following configuration: processor is an 11th Gen Intel(R) Core(TM) i7-1165G7 @ 2.80GHz, 1690 Mhz, 4 Core(s), with 16.0 GB (15.7 GB usable) RAM and 64-bit Windows 10 Enterprise operating system. We see that the fully non-parametric model performs reasonably faster than other kernels and it is to be noted that the computational time incurred using the Matérn kernel is very high with respect to all the others and thus practically infeasible to use especially when we need predictions in real time. From the point of view of accuracy of predictions and ease/quickness of computation, we put forward the full non-parametric model, with the SQE-looking kernel, as adequate in undertaking learning given a real-world data situation as the one considered in this paper.

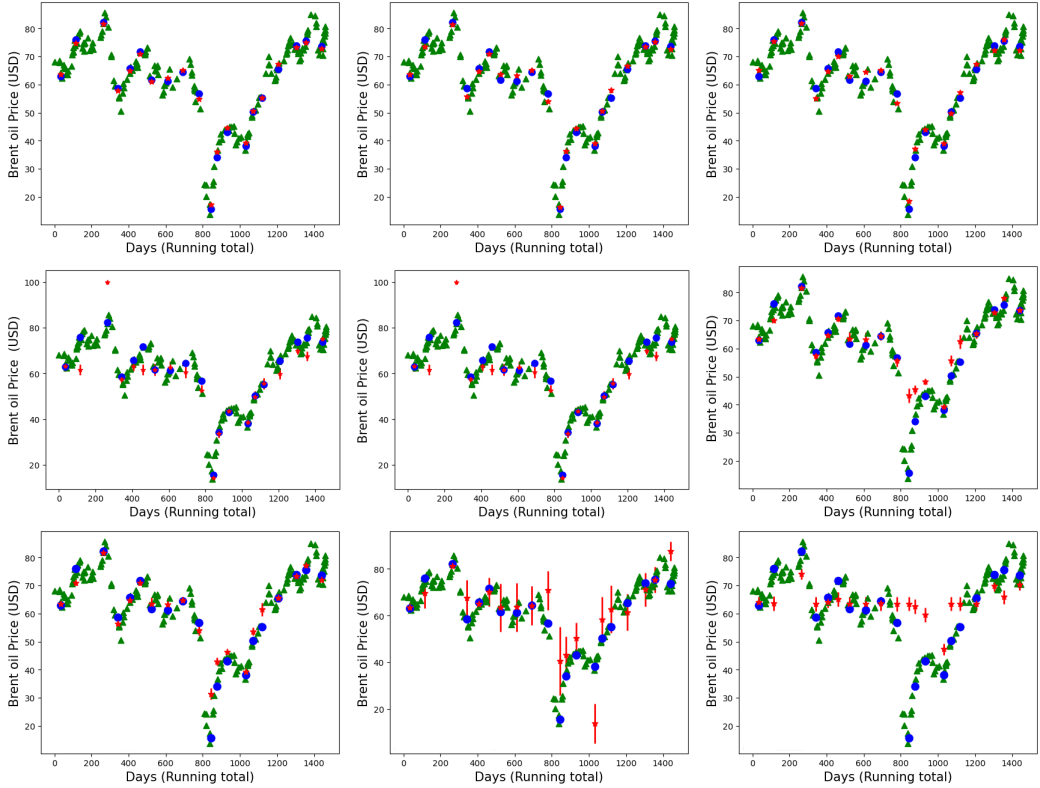

Figure 9: Figure displaying prediction of the oil price output at each of 20 identified test inputs (in red), undertaken following the learning of the temporal variation of the price made using the Brent crude oil price data. 2.5 times the predicted standard deviation of the output is added to either side of the predicted mean, as an error bar (in red). Means of the predicted outputs are presented in red stars; true values of the output are in blue circles; and training points are in green triangles. In the left panel of the top row, we depict predictions made using the fully non-parametric model with the SQE-looking kernel. The middle panel in this row depicts predictions with the fully non-parametric model using Matérn-looking kernel and the top right panel presents predictions using the non-parametric model with an SQE-looking kernel, given noisy oil price data. In the middle row, the left panel shows the predictions made using the non-stationary version of the fully non-parametric model, (where the non-stationary equivalence is constructed using the non-parametric model with an SQE-looking kernel). The middle panel in this row presents predictions made using the Blended model, while the right panel of the middle row shows predictions made using the single GP layer based model with a stationary SQE kernel. In the bottom row, the left panel shows predictions made using a single GP layer based model with stationary Matérn kernel, while the middle panel depicts predictions made using the non-stationary kernel from Remes et al. (2017) and the right panel shows predictions made using the non-stationary kernel of Paciorek and Schervish (2003).

## 5.8 DNN

We have also implemented Deep Neural Networks (DNN) with varying architectural details, to predict oil prices at the same test inputs, as have been used for prediction purposes when GP-based models were used. Here the DNN architecture is made to vary in: the number of hidden layers and in the number of neurons in each hidden layer. We make these predictions using Tensorflow, following the code that is available in the official tutorial site of Tensorflow (https://www.tensorflow.org/tutorials/keras/regression). We have used the sigmoid activation function for these DNNs along with a learning rate of 0.01. The learning rate and the epoch value are chosen in a way that the loss functions have converged.

The purpose of this experiment is to show the sensitivity of the predictions on DNN architecture. The objective is not to identify the architecture that yields the best predictions at the test inputs, but to produce corroboration of the high sensitivity of predictions to a few architectural markers of DNNs without getting insights on explainability on the quality of predictions. We present predictions obtained from three DNN models to illustrate this whose specifications are as follows.

1. NN-1: DNN with two hidden layers, first and second hidden layer consists of 250 and 184 neurons respectively; 20000 epoch is used.

2. NN-2: DNN with three hidden layers, each hidden layer has 64 neurons; 30000 epoch is used.

3. NN-3: DNN with three hidden layers, with the first, second and third hidden layer comprised of 64, 184 and 64 neurons respectively; 20000 epoch is used.

In Figure 10, the panes in the left, middle and right display results obtained using NN-1, NN-2 and NN-3 respectively. The top row depicts predictions at test inputs; in the middle row the loss function is plotted; and in the bottom row, we show results from the last 2000 epochs, to illustrate the convergence of loss. To compute the value of loss at convergence, we have discarded the first 10000. From the remaining epochs, we compute 95% confidence interval (CI) and the mean of the loss.

The MSE and 95% CI of the loss - for NN-1, NN-2 and NN-3 – respectively are 48.0624 and $[3.4144, 4.0294]$; 4.5236 and $[0.8189, 1.241]$; 320.5724 and $[10.7665, 10.7679]$. We see that a little perturbation in the architecture causes huge difference in the performance of prediction and as well as the convergence of loss. For example, NN-2 and NN-3 have same number of hidden layers, same number of neurons in the first and third hidden layer except the number of neurons in the middle hidden layer are different, have used same optimiser with same activation function along with same learning rate; yet we see the loss for NN-3 is huge with respect to NN-2 and the loss for NN-3 has not converged. We also do not get any insight on the explainability for this.

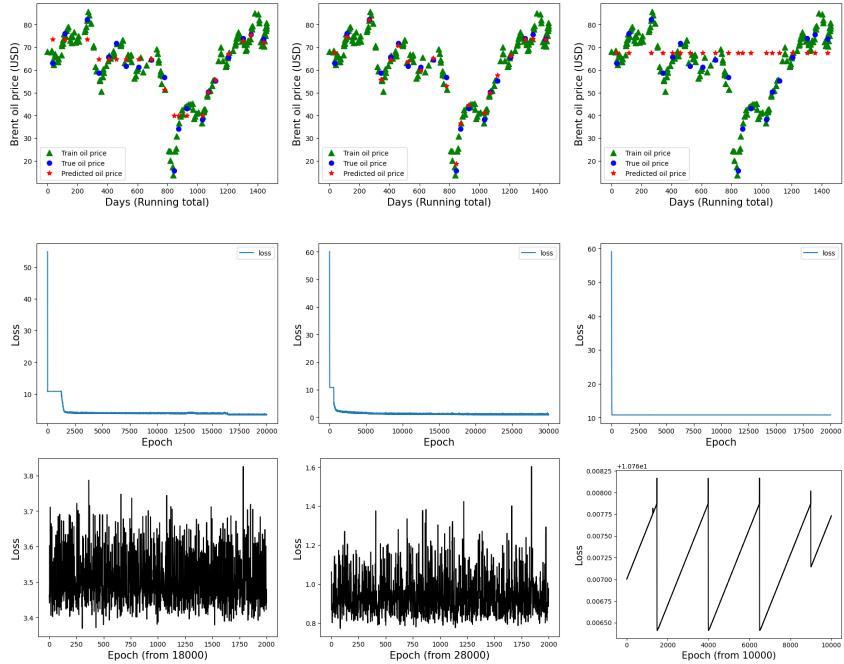

Figure 10: Columns in the left, middle and right depict results pertaining to NN-1, NN-2 and NN-3 respectively. The top row shows predictions at test inputs; in the middle row the loss function is plotted; and the bottom row presents results from the last 2000 epochs to illustrate the convergence of loss.

# 6 Conclusions

In this paper we have addressed the common problem of the need to learn a function that links an input variable to an output, given training data that manifests heterogeneities in its correlation structure. Here we have resorted to the result that sampling multiple times from a GP yields the same on the average, as achieved by sampling once from each of multiple GPs that are distinguished from each other by hyperparameters of the covariance kernel, which are updated at the equilibrium stage of the inferential exercise used to learn said hyperparameters. In such a situation, the kernel hyperparameters can be modelled as dependent on that parameter, variation of which induces a new sample function to be drawn from the relevant GP. Such a parameter is the iteration index, where a new sample function is generated at every new iteration, s.t. setting a kernel hyperparameter as given differently at each pair of input values, can be replaced by considering the hyperparameter as given differently at each iteration. In this model, we can address heterogeneity in the empirical correlation structure by learning a single hyperparameter per dimension of the input vector $\boldsymbol{X}$, for each of the two layers of GPs that we use in the learning strategy. Importantly, we prove sufficiency of stationarity of the GPs that comprise the inner layer, thereby proving sufficiency of two layers of GPs. The hyperparameters of the covariance kernel used for specification of any inner layer GP, are then constants that we learn given the lookback data which is constructed to comprise pairs of index of one – amongst an identified number of – iterations, and the hyperparameter value that was current in that iteration. Within this strategy, we have learnt the temporal variation of Brent Crude Oil prices, and predicted prices at test times. An application of this strategy to the learning of a scalar-valued multivariate function is underway, as is the learning of a matrix-valued function that takes a vector as its input.

## Acknowledgments and Disclosure of Funding

GR is funded by an EPSRC DTP studentship.

## Appendix A. Additional results

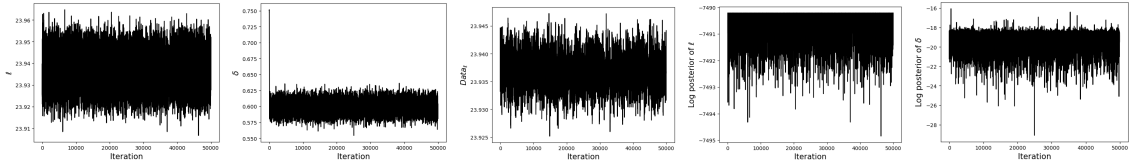

Figure 11: Results of inference made on hyperparameters of the non-parametric model with SQE-looking kernel, using Brent crude oil price data. The panels from left depict the traces of: $\ell$; $\delta$; predicted $\ell$; and logarithm of the joint posterior of $\ell$ and $\delta$, given data $D$.

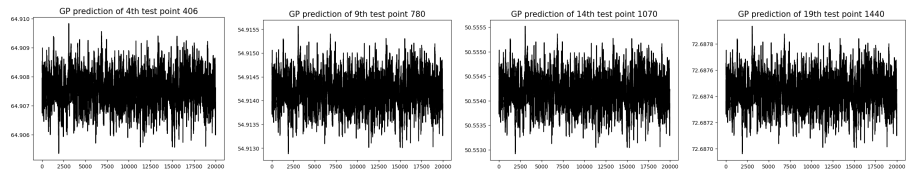

Figure 12: Traces of the output (Brent crude oil price) predicted at every fifth of the 20 selected test inputs. obtained from using the non-parametric model using SQE-looking kernel.

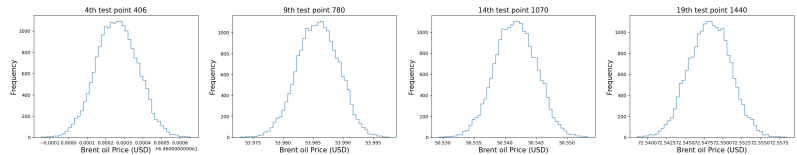

Figure 13: Histograms of outputs predicted at the test points (every fifth test input), where the predictions are obtained using non-parametric model with the Matérn-looking kernel.

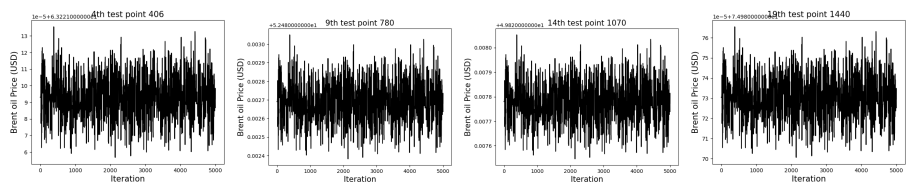

Figure 14: Traces of mean outputs predicted at the test points (every fifth test input), where the predictions are obtained using the Blended model.

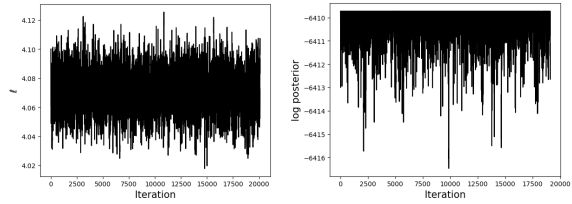

Figure 15: Traces of learnt length scale parameter $\ell$ (left) and logarithm of posterior *pdf* of $\ell$ given the training data (right), using a stationary SQE kernel.

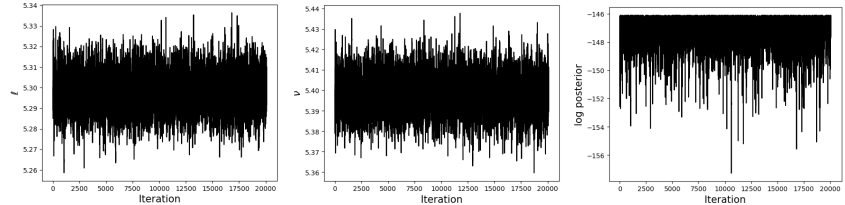

Figure 16: Traces of hyperparameters $\ell$ and $\nu$ learnt with a stationary Matérn kernel, (given the training data on Brent crude oil price data), are depicted in the left and middle panels). The right panel shows the trace of the joint log posterior of the hyperparameters given this data.

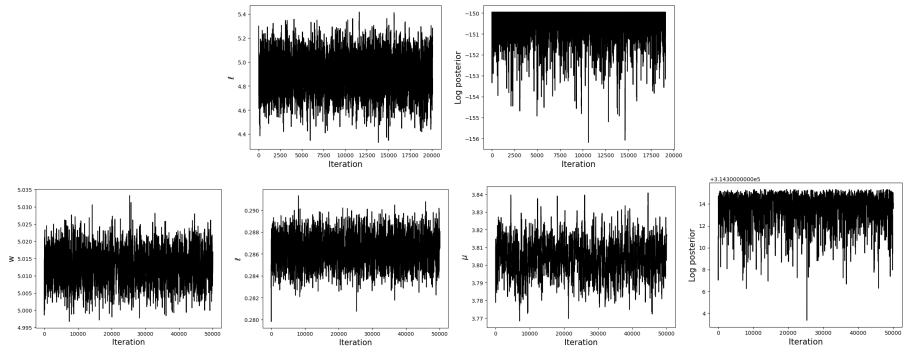

Figure 17: Panels in the top row display results of learning performed with the non-stationary kernel developed by Paciorek and Schervish (2003), given the Brent crude oil price data. Trace of the length scale hyperparameter $\ell$ of the Matérn GP prior - that is assigned a fixed $\nu = 5.3979$ - is displayed in the top left panel, while logarithm of the posterior density of this hyperparameter is on the right. The bottom row presents results of the learning performed with the non-stationary spectral kernel developed by Remes et al. (2017). Traces of the parameters $w$, $\ell$ and $\mu$ are displayed from the bottom left, while that of the joint log posterior in to the right.

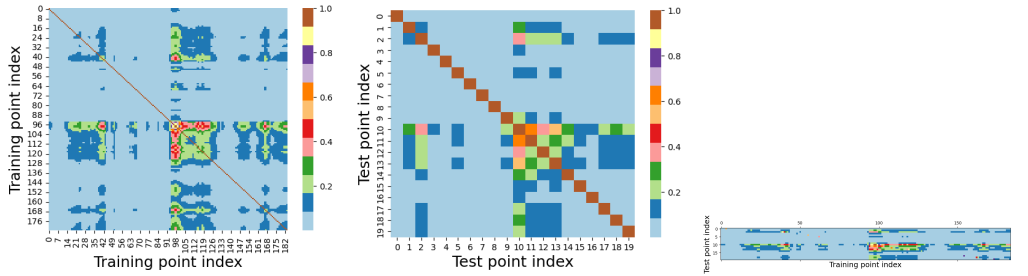

Figure 18: The left-most plot presents a heat map of the correlation matrix, any element of which is the inter-output correlation in the blended model, where each correlation element is between outputs realised at a pair of design inputs. The heat map in the middle plot represents the correlation computed at inputs pairs, where each of the pair is a test input. The right-most plot depicts the heat map of the correlation matrix, an element of which is the correlation between an output realised that is predicted at a test input, and another that is realised at a design time.

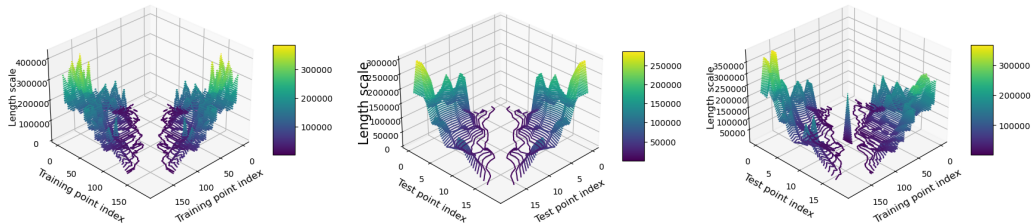

Figure 19: Similar to Figure 18, except this plots depicts the contour plots of length scales at which an inter-output correlation is kernel-parametrised by the kernel of the blended model.